\pdfoutput=1

\pdfoutput=1

\documentclass[11pt]{article}

\usepackage[]{acl}

\usepackage{times}
\usepackage{latexsym}
\usepackage[capitalize,noabbrev]{cleveref}
\usepackage{afterpage}

\usepackage[T1]{fontenc}

\usepackage[utf8]{inputenc}

\usepackage{microtype}

\usepackage{inconsolata}

\usepackage{graphicx}

\usepackage{float}

\usepackage{xcolor}

\usepackage{tcolorbox}
\usepackage{siunitx}
\usepackage{caption}
\usepackage{booktabs}
\usepackage{makecell}
\usepackage{graphicx}
\usepackage{hyperref}
\usepackage{booktabs}
\usepackage{wrapfig}

\usepackage{booktabs,multirow,array,graphicx,pifont}

\usepackage{xcolor} 
\usepackage{ifthen}
\newboolean{showcomments}
\setboolean{showcomments}{true} 
\newcommand{\AI}[1]{\ifthenelse{\boolean{showcomments}}{\textcolor{blue}{[AI: #1]}}{}}

\def\secref#1{\S\ref{sec:#1}}

\title{Through the LLM Looking Glass: A Socratic Probing of Donkeys, Elephants, and Markets}

\author{
  Molly Kennedy\textsuperscript{1}, Ayyoob Imani\textsuperscript{2}, Timo Spinde\textsuperscript{3}, Akiko Aizawa\textsuperscript{4}, Hinrich Schütze\textsuperscript{1} \\
  \textsuperscript{1}Ludwig-Maximilians-Universität München \\
  \textsuperscript{2}Microsoft \\
  \textsuperscript{3}University of Göttingen \\
  \textsuperscript{4}National Institute of Informatics, Tokyo \\
  \texttt{molly@cis.uni-muenchen.de} \\
  \texttt{ayyoobi@microsoft.com}
}

\newcounter{notecounter}

\newcommand{\enoteson}{\long\gdef\enote##1##2{{
\stepcounter{notecounter}
{\large\bf
\hspace{0cm}\arabic{notecounter} $<<<$ ##1: ##2
$>>>$\hspace{1cm}}}}}

\enoteson

\begin{document}
\maketitle
\begin{abstract}

Large Language Models (LLMs) are widely used for text generation, making it crucial to address potential bias. This study investigates ideological framing bias in LLM-generated articles, focusing on the subtle and subjective nature of such bias in journalistic contexts. We evaluate eight widely used LLMs on two datasets—\textsc{PoliGen} and \textsc{EconoLex}—covering political and economic discourse where framing bias is most pronounced. Beyond text generation, LLMs are increasingly used as evaluators (LLM-as-a-judge), providing feedback that can shape human judgment or inform newer model versions. Inspired by the Socratic method, we further analyze LLMs’ feedback on their own outputs to identify inconsistencies in their reasoning. Our results show that most LLMs can accurately annotate ideologically framed text, with GPT-4o achieving human-level accuracy and high agreement with human annotators. However, Socratic probing reveals that when confronted with binary comparisons, LLMs often exhibit preference toward one perspective or perceive certain viewpoints as less biased.

\end{abstract}

\section{Introduction}

\begin{figure}[!t]
    \centering
    \begin{minipage}[b]{0.537\linewidth}
        \centering
        \includegraphics[width=\linewidth]{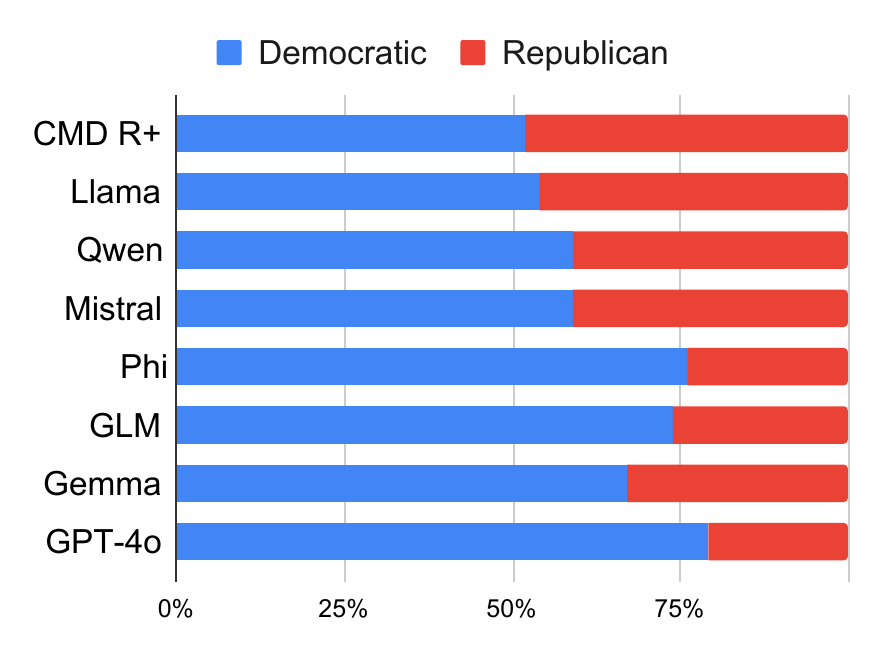}
        \label{fig:1}  
    \end{minipage}
    \hfill
    \begin{minipage}[b]{0.443\linewidth}
        \centering
        \includegraphics[width=\linewidth]{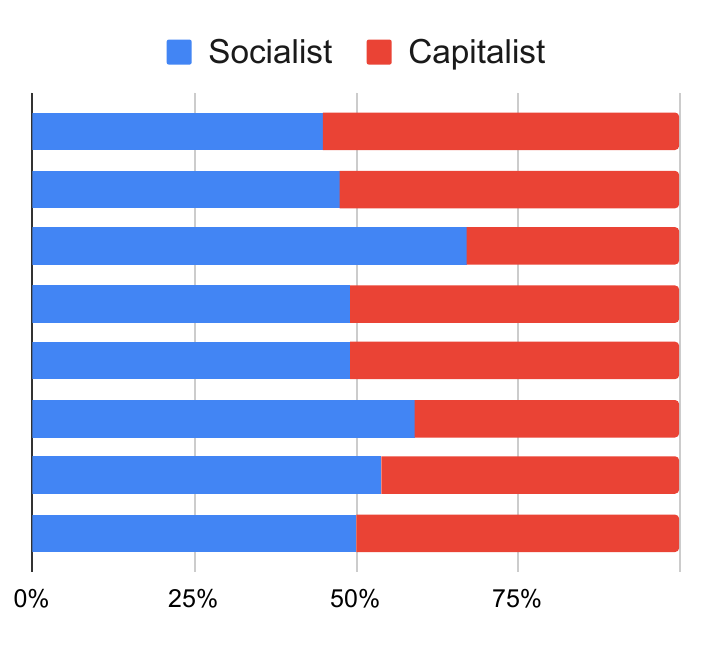}
        \label{fig:2}  
    \end{minipage}
    
    \caption{%
    We first ask each LLM to generate a document $d$ from a different perspective (e.g., ``Democratic'') and then
    ask it to assess $d$'s preference
    without including any information about the generation
    process.
    Results of this generation and self-assessment method
    are shown for eight LLMs.
        \textbf{Left:}
        LLMs prefer 
Democratic over Republican positions in self-generated articles on political topics.
        \textbf{Right:}
While there is no consistent trend for Western models,
    Chinese models prefer
Socialist over Capitalist positions in self-generated articles on economical titles.%
    }
    \label{fig:combined}
\end{figure}

Media bias, especially in the form of ideological framing, plays a powerful role in shaping public opinion \citep{mokhberian2020moral}. News articles on the same event can highlight different values, priorities, or causal narratives, subtly steering how readers interpret political or economic issues \citep{pastorino2024decoding}. Detecting such framing bias is challenging: it is often subtle, context-dependent, and subject to the annotator’s own perspective \citep{wang2025media}. 

At the same time, large language models (LLMs) are increasingly used not only to generate text but also to evaluate it \citep{li2024llms}
. This dual role raises an important question: when LLMs generate media-like content, do they reproduce ideological leanings, and when they judge such content, can their evaluations be trusted? Existing research has probed bias in pretrained models \citep{weidinger2022taxonomy, Rishi2021Foundation, lin-etal-2024-indivec, bang-etal-2024-measuring} and examined model leanings in political or social contexts \citep{sheng-etal-2021-societal, lin2024investigating, trhlik2024quantifying, fang2024bias, hernandes2024llmsleftrightcenter}. Other work has explored “LLM-as-a-judge” for diverse evaluation tasks \citep{zheng2023judging, liu2023g, kim2023prometheus, kim2024prometheus}. Yet three challenges remain: human annotation of bias is expensive and subjective \citep{lin2023data, spinde_thinkIsBiased}; fine-tuned classifiers capture only limited framing nuances \citep{spinde2022neural, Horych2025}; and model-based evaluation itself risks inheriting hidden ideological priors \citep{potter2024hidden, gu2024survey, gu2025alignment}.

To address these challenges, we present a study of \emph{media bias in LLM-generated text}. We assess eight widely used open and commercial LLMs across political and economic domains in three complementary roles. First, we collected 1,000 political topics and 1,000 economic news headlines. All topics and titles were human-verified to ensure relevance (i.e., appropriateness for the task) and non-redundancy (i.e., sufficiently distinct from one another), forming the \textsc{PoliGen} (political) and \textsc{EconoLex} (economic) datasets. Using these datasets, we then prompted the LLMs to generate articles from different ideological perspectives. We then benchmark these outputs through human annotation and compare them to model-based annotation, quantifying accuracy and agreement. Finally, we introduce \emph{Socratic probing} as a complementary diagnostic, where it is used to reveal contradictions and latent leanings that are not captured by agreement metrics alone. The Socratic method has been used in education and the sciences to stimulate critical thinking \citep{ho2023thinking, theartofsocraticquestioning2023}, uncover hidden assumptions \citep{wang2024socreval, socraticquestioningmultimodal2025}, and guide systematic self-correction \citep{he_selfcorrection2024, socraticRL2025}. We use it to provide us with an additional lens for analyzing bias. 

Our findings show that GPT-4o achieves near-human agreement with annotators in ideological labeling tasks, with accuracy of 97\% and Cohen’s $\kappa=0.90$, closely mirroring human consensus (91\%). By contrast, Command-R performed poorly, with only 66\% alignment, 48\% accuracy, and very weak agreement ($\kappa=0.14$). Other open-weight models such as Llama, Qwen, and Mistral fell in a moderate range, reflecting partial but imperfect accuracy and agreement with human judgments.
Socratic probing further reveals inconsistencies in self-assessment, showing that GPT-4o, despite its high accuracy in identifying article framing, judged capitalist articles as less biased in 94\% of binary economic comparisons.
Other systematic preferences observed in binary comparisons include: in political topics, GPT-4o and Phi favor Democratic framings (e.g., GPT-4o 79\% Democrat vs.\ 21\% Republican), while in economic topics, Chinese-developed models such as Qwen (67\% Socialist vs.\ 33\% Capitalist) and GLM (55\% vs.\ 45\%) consistently favor socialism. Together, these results underscore both the promise and the limitations of LLMs as evaluators of media bias, and point toward the need for layered evaluation methods that combine human annotation, model judgment, and deeper probing.

We will publicly share all code and data. \footnote{Available at:  
\url{https://github.com/MolKennedy95/LLM_bias_analysis}.

}

\section{Related Work}\label{sec:RW}

\subsection*{Political and Ideological Bias in Language Models}

Recent studies have shown that large language models (LLMs) exhibit various forms of bias, including political, gender, and religious biases \cite{bender2021stochastic,weidinger2022taxonomy,Rishi2021Foundation,cho2020investigating,winoqueer2023benchmark,persistent2021bias,beyond2024labelbias}. 


Most work on LLM political behavior evaluates models in \emph{classification}-like settings (e.g., predicting ideology labels or answering political questionnaires) \citep{rottger2024political, elbouanani2025analyzing, haller2025leveraging}.
Building on this paradigm, more recent work has begun to evaluate prompted LLMs for media-bias detection across model families \citep{maab2024media, faulborn2025only}.

Several studies demonstrate that small changes in prompts can significantly shift a model's political leaning \cite{bang-etal-2024-measuring}, while instruction-tuned LLMs tend to reinforce users’ pre-existing ideologies \cite{hernandes2024llmsleftrightcenter}. \citet{lin-etal-2024-indivec} introduce \textit{IndiVec}, a method for capturing subtle ideological shifts resulting from minor lexical variations, and \citet{trhlik-stenetorp-2024-quantifying} find that instruction-tuned models more strongly align with ideological narratives than their base counterparts.

Beyond politics, other dimensions of social bias have also been studied. For example, \citet{instructed2023bias} find that instruction tuning can amplify pre-existing tendencies in LLMs, while \citet{persistent2021bias} highlight systematic anti-Muslim bias. Broader efforts such as \citep{bold2024biasdataset} aim to quantify open-ended bias through dataset development and targeted metrics.

Despite these advances, existing approaches often suffer from several limitations: (1) a predominant focus on media bias detection using encoder-only models without fully utilizing the introspective capabilities of generative LLMs \cite{Spinde2023MediaBias,Horych2025}, (2) an emphasis on general fairness rather than specific ideological or political leanings \cite{motoki2024more,gehman-etal-2020-realtoxicityprompts}, (3) evaluations restricted to narrow topics, which may miss more nuanced ideological variance \cite{bang-etal-2024-measuring,hernandes2024llmsleftrightcenter,trhlik-stenetorp-2024-quantifying}, and (4) a lack of self-contained mechanisms that enable models to critically examine their own outputs \cite{lin-etal-2024-indivec}.

\section{Media Bias and Framing}
Media bias refers to systematic patterns in how news outlets frame events, select facts, and use language to privilege certain perspectives over others \cite{entman2007framing, mcquail2020mcquail}. Unlike demographic or representational biases embedded within language models, media bias is inherently tied to the socio-political context of news reporting and is observable only within the outputs of specific outlets. It is expressed not only through explicit opinions or partisan narratives but also through more subtle mechanisms such as the prioritization of certain themes, the use of emotionally charged language, or the omission of alternative perspectives.

A clear illustration of media bias can be seen in the coverage of the Respect for Marriage Act \cite{vallejo2023connecting}. In this case, left-leaning outlets framed the legislation as a historic win for LGBTQ+ rights, emphasizing bipartisan support and presenting celebratory imagery. In contrast, right-leaning outlets framed the event around political contestation and judicial overreach, focusing on potential constitutional challenges. This example highlights how media bias often operates through framing choices rather than overt factual distortions, subtly guiding audience interpretations.

In our study, we build upon this understanding of framing bias by evaluating the capacity of large language models to generate ideologically framed outputs that mirror such nuanced forms of bias. Although the models were prompted to produce articles from a range of ideological perspectives—including Democratic, Republican, Socialist, Capitalist, and Neutral—we illustrate the phenomenon here using an example contrasting Socialist and Capitalist framings (see Figure~\ref{fig:ideological-framing}). These examples are drawn from our experiments, which are detailed in Section~\ref{sec:experiment-setup}
. Given the same headline, \textit{"How to Focus in a Manic Market,"} the Socialist-framed article critiques the manic market as a symptom of capitalist systems prioritizing speculative profits over societal welfare, adopting a cautionary tone and highlighting systemic risks to ordinary people. In contrast, the Capitalist-framed article embraces market volatility as a source of opportunity, framing it as an inherent feature of dynamic economies and encouraging investors to capitalize on the chaos. These examples demonstrate how LLM-generated outputs can reflect framing bias through shifts in tone, emphasis, and value prioritization, paralleling how media outlets present the same events through different ideological lenses (see Figure~\ref{fig:ideological-framing}).

\begin{figure*}[t]
\centering
\begin{minipage}[t]{0.48\textwidth}
\begin{tcolorbox}[colframe=red!70!black, colback=white, sharp corners, boxrule=0.8pt, left=1mm, right=1mm, top=1mm, bottom=1mm]
\raggedright

\textbf{Prompt:} Write an article on the following title from the perspective
of a Socialist journalist.

\textbf{Title: How to Focus in a Manic Market}

\textbf{LLM-generated article:}
[...] In today's global economy, financial markets often resemble a \textbf{rollercoaster ride}, characterized by \textbf{extreme volatility and unpredictability. This manic market} behavior can be disorienting, especially for those who are not part of the financial elite. From a perspective centered on economic justice, it is crucial to understand the underlying \textbf{forces driving these market fluctuations} and to focus on strategies that prioritize collective well-being over individual profit.

[...] The manic market is a product of a system that prioritizes \textbf{profit maximization and speculative investment} over stable, long-term growth. This encourages \textbf{risky behavior}, as financial institutions and investors \textbf{chase short-term gains}, often \textbf{at the expense of economic stability}. The result is \textbf{a market prone to bubbles and crashes}, leaving ordinary people vulnerable to economic shocks.
\end{tcolorbox}
\end{minipage}
\hspace{0.02\textwidth}
\begin{minipage}[t]{0.48\textwidth}
\begin{tcolorbox}[colframe=blue!70!black, colback=white, sharp corners, boxrule=0.8pt, left=1mm, right=1mm, top=1mm, bottom=1mm]
\raggedright

\textbf{Prompt:} Write an article on the following title from the perspective
of a Capitalist journalist.

\textbf{Title: How to Focus in a Manic Market}

\textbf{LLM-generated article:}
[...] In the ever-evolving landscape of global finance, markets can often swing from calm to chaotic with startling speed. The recent surge in market volatility has left many investors and business leaders grappling with uncertainty. However, for those embracing market principles, \textbf{these manic markets present not just challenges, but opportunities}. Here’s how to maintain focus and \textbf{capitalize on the chaos}.

First and foremost, understanding that \textbf{markets are inherently dynamic} is crucial. \textbf{Volatility is not an anomaly but a natural part of the economic cycle}. Businesses \textbf{thrive on change}, and a manic market is a \textbf{testament to the fluidity of supply and demand. Embrace this dynamism as a catalyst for innovation and growth.}
\end{tcolorbox}
\end{minipage}
\caption{LLM-generated articles with contrasting ideological framings. Both articles share the same headline but differ in framing emphasis. The left article critiques market instability as systemic failure, focusing on collective welfare and economic risks. The right article frames volatility as an opportunity for growth, emphasizing entrepreneurial resilience. Key phrases illustrating framing differences are bolded for clarity. Text excerpts are abridged for illustration.}
\label{fig:ideological-framing}
\end{figure*}



\section{Datasets}

In this study, we introduce two datasets, which human reviewers have verified for quality and relevance to economic and political discourse. \textsc{PoliGen} contains 1,000 topics across ten political categories, while \textsc{EconoLex} comprises 1,048 real-world economic news titles. The datasets are summarized in \Cref{tab:dataset_categories}.

\paragraph{\textsc{PoliGen} Political Topics}
\label{sec:poligen}
\textsc{PoliGen} is generated using GPT-4o. The initial prompting produced ten political categories, then topics were generated under each category relevant to the U.S. election. To ensure quality and avoid redundancy, the generated topics were manually reviewed domain experts on the author team. Specifically, duplicate or overly similar topics were removed, and the final selection maintained a balanced representation of political themes across diverse ideological viewpoints.

\paragraph{\textsc{EconoLex} Economical Titles}
\label{sec:econolex}
\textsc{EconoLex} comprises 1,048 economic news titles from the publicly available FNSPID datasets \cite{dong2024fnspid}. Titles were selected for their potential to support economic analysis, with human verification ensuring they reflect differing socialist and capitalist perspectives. Articles focused solely on financial metrics (e.g., stock performance, ETFs, earnings reports) were excluded in favor of those discussing economic policies and financial decisions open to ideological interpretation.

\section{Methodology}
\label{sec:methodology}

We aim to assess the extent of bias in large language models (LLMs). Our methodology combines three components. First, we prompted eight SOTA LLMs to generate articles on political (\textsc{PoliGen}, \secref{poligen}) and economic (\textsc{EconoLex}, \secref{econolex}) topics from contrasting ideological perspectives, ensuring balanced coverage across Democratic, Republican, and neutral (political) as well as Socialist, Capitalist, and neutral (economic) framings. We then evaluated the ideological alignment of these outputs using both human and model-based annotation, enabling direct comparison of model fidelity to prompts and human–model agreement (quantified via Cohen’s $\kappa$). Finally, we conducted Socratic probing, asking models to compare the three ideologically framed articles for each topic and each dataset, once to indicate their \textit{preferred} article and once to identify the \textit{least biased}. This approach reveals implicit ideological leanings and exposes contradictions in model reasoning. Together, this multi-part design captures bias from three angles: generation, annotation, and Socratic probing, providing an assessment of LLM ideological behavior.

\section{Experiment Setup}
\label{sec:experiment-setup}
We selected eight  SOTA LLMs from different families for a comprehensive analysis. Table~\ref{tab:llm_list} provides their details.

\begin{table}[h]
    \centering
    \scriptsize
    \caption{List of Large Language Models (LLMs) Used in the Experiment}
    \label{tab:llm_list}
    \begin{tabular}{lll}   
        \toprule
        \textbf{Model Name} & \textbf{Model Details}  & \textbf{Developer} \\
        \midrule
        CMD R+ & c4ai-command-r7b-12-2024 & Cohere AI \\ 
        Llama & Llama-3.1-8B-Instruct & Meta \\ 
        Qwen & Qwen2.5-7B-Instruct & Alibaba/Qwen \\ 
        Mistral & Ministral-8B-Instruct-2410 & Mistral AI \\ 
        Phi & Phi-4-14B & Microsoft \\ 
        GLM & glm-4-9b-chat & THUDM \\ 
        Gemma & gemma-2-9b-it & Google \\ 
        GPT-4o & chatgpt-4o-latest & OpenAI \\ 
        \bottomrule
    \end{tabular}
    
\end{table}


\subsection{Article Generation} 
\label{stage1}

We generated articles across distinct ideological perspectives to construct the \textsc{PoliGen} and \textsc{EconoLex} datasets. For \textsc{PoliGen}, models alternated between Democratic, Republican, and neutral perspectives; for \textsc{EconoLex}, articles were written from Socialist, Capitalist, and neutral perspectives. To ensure balance and systematic coverage, we employed predefined combinations of system and user prompts (Tables~\ref{table:article_prompt_combos} and \ref{table:article_prompt_combos_real}).

For each dataset, we used the default generation parameters of the respective models, following their chat template structures with both system messages and user inputs. Articles were generated with a maximum length of 512 tokens.

\subsection{Human and Model Annotation Study} 
\label{stage2}

To assess whether model outputs aligned with their intended ideological prompts, we conducted a joint human–model annotation study on a subset of 1,008 articles sampled from \textsc{EconoLex} (\secref{econolex}). Specifically, we first selected 30 topics and then annotated articles generated by each model within those topics. We sampled an equal number of articles from each model, yielding a balanced per-model evaluation set. Each selected article was independently annotated by three annotators.

Articles were categorized as Socialist, Capitalist, or Neutral following detailed guidelines grounded in framing theory (Appendix~\ref{appendix:annotators}). Annotators focused on thematic and value-based cues (e.g., equality, fairness, efficiency) and tonal markers, rather than relying on explicit ideological keywords, which were masked to prevent lexical bias. Examples defining each category were drawn from authentic news sources with known leanings, anchoring judgments in real-world framing conventions.

Thirteen annotators contributed to this study, they were not informed of the prompts used to generate the articles, ensuring independence of judgment.

In parallel, we asked the same models to annotate the identical set of articles under the same categorical framework (see Appendix Figure~\ref{fig:model-prompt} for the full prompt). This setup enabled two complementary assessments: (i) an external benchmark of how accurately models generated the requested perspectives, and (ii) a direct comparison of human and model annotations, quantified in later sections using Cohen’s $\kappa$ to measure agreement (see Section~\ref{sec:results}).

\subsection{Preference and Least-Bias Evaluation}
\label{stage3}

To probe models’ evaluative biases, we asked them to compare the articles generated in \Cref{stage1}. For each topic, models were presented with either two (excluding neutral) or three (including neutral) articles written from different ideological perspectives. They were instructed to identify either (i) their \textit{preferred} article or (ii) the article they judged \textit{least biased} (see \cref{preference_prompts} for prompt details).

Importantly, models were not informed of the ideological perspective associated with each article, ensuring blindness to the \Cref{stage1} generation process. A consistent preference for one perspective was interpreted as evidence of ideological leaning, while least-bias judgments shed light on how models evaluate neutrality.

To minimize confounds, the candidate articles were randomly shuffled and assigned only numerical identifiers. Models were instructed to respond strictly by number, and outputs were constrained to a maximum of 10 tokens under greedy decoding to ensure deterministic responses. Responses were parsed with regular expressions, and any unprocessable outputs (fewer than one percent across all models) were skipped.

\begin{table}
  \sisetup{group-minimum-digits = 4}
  \centering
  \scriptsize
  \caption{Combinations of system prompt and user prompt
  that we used 
to generate articles on \textsc{PoliGen}
  topics. The language model subsequently evaluated both
  biased and neutral articles to determine preference.}
  \label{table:article_prompt_combos}
  
  \begin{tabular}{p{3cm} p{3.3cm}}  
    \toprule
    System prompt & User prompt \\
    \midrule
    "You are a helpful AI assistant." & "Write a \{Democratic/Republican\} article on topic <topic>." \\
    \midrule
    "You are a helpful AI assistant." & "Write an article on topic <topic>." \\
    \midrule
    "You are a \{Democratic/Republican\} journalist." & "Write an article on topic <topic>." \\

    \bottomrule
  \end{tabular}
  \label{table:generation_prompts}
\end{table}

\begin{table}[h]
  \centering
  \scriptsize
  \caption{Combinations of system prompt and user prompt
  that we used 
to generate articles on \textsc{EconoLex}
  titles. The language model subsequently evaluated both
  biased and neutral articles to determine preference.}
  \label{table:article_prompt_combos_real}
  \begin{tabular}{p{2.8cm} p{4cm}}  
    \toprule
    \textbf{System Prompt} & \textbf{User Prompt} \\
    \midrule
    "You are a helpful AI assistant." & "Write an article on the following title from the perspective of a \{Socialist/Capitalist\} journalist. Title: <title>" \\
    \midrule
    "You are a helpful AI assistant." & "Write an article on the following title. Title: <title>" \\
    \midrule
    "You are a \{Socialist/Capitalist\} journalist." & "Write an article on the following title. Title: <title>" \\
    \bottomrule
  \end{tabular}
  \label{table:generation_prompts2}
\end{table}

\begin{table}[ht]
\centering
\resizebox{\linewidth}{!}{%
\begin{tabular}{lcccc}
\hline
\textbf{Model} & \textbf{Human Accuracy} & \textbf{Model Accuracy} & \textbf{Cohen's $\kappa$} & \textbf{Human IAA (Light's $\kappa$)} \\
\hline
CMD R+   & 66\% & 48\% & 0.14 & 0.04 \\
Llama    & 85\% & 77\% & 0.37 & 0.47 \\
Qwen     & 79\% & 68\% & 0.20 & \underline{0.70} \\
Mistral  & 75\% & 70\% & 0.41 & 0.46 \\
Phi      & 79\% & 76\% & \underline{0.53} & 0.25 \\
GLM      & \underline{86\%} & 67\% & 0.46 & 0.39 \\
Gemma    & 66\% & \underline{89\%} & 0.51 & 0.62 \\
GPT-4o   & \textbf{91\%} & \textbf{97\%} & \textbf{0.90} & \textbf{0.80} \\
\hline
\end{tabular}}
\caption{Comparison of human annotation accuracy, model self-annotation accuracy, human--model agreement (Cohen’s $\kappa$), and human inter-annotator agreement (Light’s $\kappa$) across eight large language models (LLMs). The highest score in each column is shown in \textbf{bold}, and the second highest is \underline{underlined}.}
\label{tab:human-model-results}
\end{table}

\begin{table*}[htbp]
\caption{Comparison of models’ \textit{least-bias outcomes} across political and economic dimensions. 
``Two Way Political'' uses row~1 of Table~\ref{table:generation_prompts}. 
``Three Way Political'' uses rows~1 and~2 of Table~\ref{table:generation_prompts}. 
``Two Way Economic'' uses row~1 of Table~\ref{table:generation_prompts2}. 
``Three Way Economic'' uses rows~1 and~2 of Table~\ref{table:generation_prompts2}. 
For each column, the largest value is shown in \textbf{bold}, and the second-largest value is \underline{underlined}.}

  \centering
  \scriptsize 
  \renewcommand{\arraystretch}{0.9} 
  \setlength{\tabcolsep}{3pt} 
  \makebox[\textwidth][c]{
    \begin{tabular}{c|cc|ccc|cc|ccc}
      \toprule
      & \multicolumn{2}{c|}{Two Way Bias} & \multicolumn{3}{c|}{Three Way Bias} & \multicolumn{2}{c|}{Two Way Economic} & \multicolumn{3}{c}{Three Way Economic} \\
      \cmidrule(lr){2-3} \cmidrule(lr){4-6} \cmidrule(lr){7-8} \cmidrule(lr){9-11}
      Model & Democrat & Republican & Democrat & Republican & Neutral & Socialist & Capitalist & Socialist & Capitalist & Neutral \\
      \midrule
      CMD R+ & 0.50 & \textbf{0.50} & 0.31 & \textbf{0.33} & 0.36 & \underline{0.56} & 0.44 & \underline{0.32} & \underline{0.20} & 0.48 \\
      Llama & 0.52 & \underline{0.48} & \underline{0.43} & \underline{0.14} & 0.43 & 0.48 & 0.50 & 0.08 & \textbf{0.26} & 0.65 \\
      Qwen & 0.80 & 0.20 & 0.36 & 0.06 & 0.58 & 0.38 & \underline{0.62} & 0.09 & 0.13 & 0.78 \\
      Mistral & 0.73 & 0.27 & \textbf{0.47} & 0.03 & 0.50 & \textbf{0.73} & 0.27 & \textbf{0.47} & 0.03 & 0.50 \\
      Phi & \textbf{0.91} & 0.09 & 0.41 & 0.04 & 0.60 & 0.40 & 0.60 & 0.11 & 0.17 & 0.72 \\
      GLM & \underline{0.81} & 0.19 & \underline{0.43} & 0.03 & 0.53 & 0.52 & 0.48 & 0.12 & 0.15 & 0.73 \\
      Gemma & 0.79 & 0.21 & 0.28 & 0.05 & \underline{0.66} & 0.43 & 0.57 & 0.06 & 0.14 & \underline{0.81} \\
      GPT-4o & 0.53 & 0.47 & 0.01 & 0.01 & \textbf{0.98} & 0.06 & \textbf{0.94} & 0.00 & 0.02 & \textbf{0.98} \\

      \bottomrule
    \end{tabular}
  }
  \label{tab:comparison_bias}
\end{table*}

\begin{table*}[htbp]
\caption{Comparison of models’ \textit{preference outcomes} across political and economic dimensions. 
``Two Way Political'' uses row~1 of Table~\ref{table:generation_prompts}. 
``Three Way Political'' uses rows~1 and~2 of Table~\ref{table:generation_prompts}. 
``Two Way Economic'' uses row~1 of Table~\ref{table:generation_prompts2}. 
``Three Way Economic'' uses rows~1 and~2 of Table~\ref{table:generation_prompts2}. 
For each column, the largest value is shown in \textbf{bold}, and the second-largest value is \underline{underlined}.}

  \centering
  \scriptsize 
  \renewcommand{\arraystretch}{0.9} 
  \setlength{\tabcolsep}{3pt} 
  \makebox[\textwidth][c]{
    \begin{tabular}{c|cc|ccc|cc|ccc}
      \toprule
      & \multicolumn{2}{c|}{Two Way Political} & \multicolumn{3}{c|}{Three Way Political} & \multicolumn{2}{c|}{Two Way Economic} & \multicolumn{3}{c}{Three Way Economic} \\
      \cmidrule(lr){2-3} \cmidrule(lr){4-6} \cmidrule(lr){7-8} \cmidrule(lr){9-11}
      Model & Democrat & Republican & Democrat & Republican & Neutral & Socialist & Capitalist & Socialist & Capitalist & Neutral \\
      \midrule
      CMD R+ & 0.52 & \textbf{0.48} & 0.35 & \textbf{0.33} & 0.32 & 0.45 & \textbf{0.55} & 0.26 & \underline{0.34} & \underline{0.40} \\
      Llama & 0.54 & \underline{0.46} & \underline{0.49} & 0.17 & 0.34 & 0.47 & \underline{0.52} & 0.27 & \textbf{0.39} & 0.33 \\
      Qwen & 0.59 & 0.41 & 0.45 & 0.21 & 0.34 & \textbf{0.67} & 0.33 & 0.35 & 0.27 & 0.38 \\
      Mistral & 0.59 & 0.41 & \textbf{0.50} & \underline{0.22} & 0.28 & 0.49 & 0.51 & \underline{0.41} & 0.32 & 0.27 \\
      Phi & \underline{0.76} & 0.24 & 0.44 & 0.17 & \underline{0.39} & 0.49 & 0.51 & 0.34 & 0.33 & 0.33 \\
      GLM & 0.74 & 0.26 & 0.48 & 0.15 & 0.37 & \underline{0.59} & 0.41 & \textbf{0.45} & 0.28 & 0.27 \\
      Gemma & 0.67 & 0.33 & \underline{0.49} & 0.12 & \underline{0.39} & 0.54 & 0.46 & 0.31 & 0.29 & \underline{0.40} \\
      GPT-4o & \textbf{0.79} & 0.21 & 0.23 & 0.04 & \textbf{0.73} & 0.50 & 0.50 & 0.17 & 0.25 & \textbf{0.57} \\

      \bottomrule
    \end{tabular}
  }
  \label{tab:comparison}
\end{table*}

\section{Experiment results}\label{sec:results}

\subsection{Model--Human Agreement on Ideological Annotation}

To further probe how reliably LLMs can serve as annotators, we compared model predictions with human annotations across the same 1{,}000 articles. Following the human study, models were asked to label each article as Socialist, Capitalist, or Neutral, and agreement with human majority labels was quantified using Cohen’s $\kappa$ (Table~\ref{tab:human-model-results}). We additionally report human inter-annotator agreement (Light’s $\kappa$) for each model-specific subset to contextualize model--human alignment.

Results reveal substantial variation in model reliability. GPT-4o achieved near-human performance, with $\kappa=0.90$ and the highest model accuracy (97\%), closely matching human consensus accuracy (91\%); its subset also exhibited the strongest human agreement (Light’s $\kappa=0.80$). In contrast, Command-R underperformed (48\% model accuracy; $\kappa=0.14$) and was associated with very low human agreement (Light’s $\kappa=0.04$), suggesting more ambiguous or inconsistently framed outputs. Other models fall in a moderate range of human--model agreement ($\kappa=0.37$--$0.53$), but their human IAA varies widely (Light’s $\kappa=0.25$--$0.70$), indicating that differences in model--human alignment partly reflect differences in labeling difficulty across model generations (e.g., Qwen: Light’s $\kappa=0.70$ vs.\ $\kappa=0.20$).

While these results suggest that GPT-4o approximates human annotator quality in many respects, our subsequent analysis indicates that the model can still exhibit systematic internal biases. Accordingly, it should be used with caution, particularly in settings where it serves as a judge or source of evaluative feedback.

\subsection{Socratic Probing for Identifying Hidden Asymmetries}

To examine inconsistencies in model reasoning, we introduce a \textit{Socratic probing} framework, a two-step diagnostic procedure. The model first generates or labels articles from distinct ideological perspectives, then evaluates its own outputs without direct access to the originals. Rather than serving as correction, Socratic probing surfaces contradictions and anomalous reasoning patterns, helping identify imbalances in how LLMs assess politically or ideologically sensitive content.  

Tables~\ref{tab:comparison_bias} and~\ref{tab:comparison} summarize model behavior under this setup for political and economic topics, reporting \emph{least-bias outcomes} (selection frequencies) and \emph{preference outcomes} (proportions of preferred framings) under both \emph{two-way} and \emph{three-way} conditions (see Section~\ref{stage3}). Together, they reveal where model behavior remains balanced versus systematically asymmetric, and how including a Neutral option alters these tendencies.  

On political topics, most models exhibit asymmetries favoring Democratic framings; GPT-4o, Phi, and GLM show the strongest imbalances, while Command-R+ and Llama remain closer to parity. For economic topics, mild skew appears across models, with Qwen and GLM leaning more toward Socialist framings. These trends echo prior findings that LLMs may reflect characteristics of their training data or development context \citep{buyl2025largelanguagemodelsreflect}.  

Adding a Neutral option yields only modest shifts: GPT-4o favors Neutral responses 73\% of the time for political and 57\% for economic topics, while others change little. Economic classification is generally more balanced, perhaps reflecting broader conceptual diversity than the polarized framing of U.S. politics.  

Table~\ref{tab:comparison_bias} also illustrates why GPT-4o’s reliability merits caution. With three-way options, it identifies Neutral articles nearly perfectly (98\%), but under binary setups, it overwhelmingly selects one side (e.g., 94\% Capitalist, 53\% Democrat). Other models show analogous but differently directed biases (e.g., Phi favoring Socialist framings). CMD R+ displays the least skew across prompts and datasets, likely due less to fairness than to weaker bias understanding, as reflected in its poor annotation and human-agreement scores (Table~\ref{tab:human-model-results}).  

Overall, high performance on neutrality tasks can mask deeper asymmetries revealed through Socratic probing. While not definitive evidence of ideological bias, these findings highlight behaviors that warrant caution, especially when LLMs are used as evaluators or feedback providers.

\paragraph{Preference vs. Bias}

\begin{figure}[ht]
    \centering
    \begin{minipage}[b]{0.49\linewidth}
        \centering
        \includegraphics[width=\linewidth]{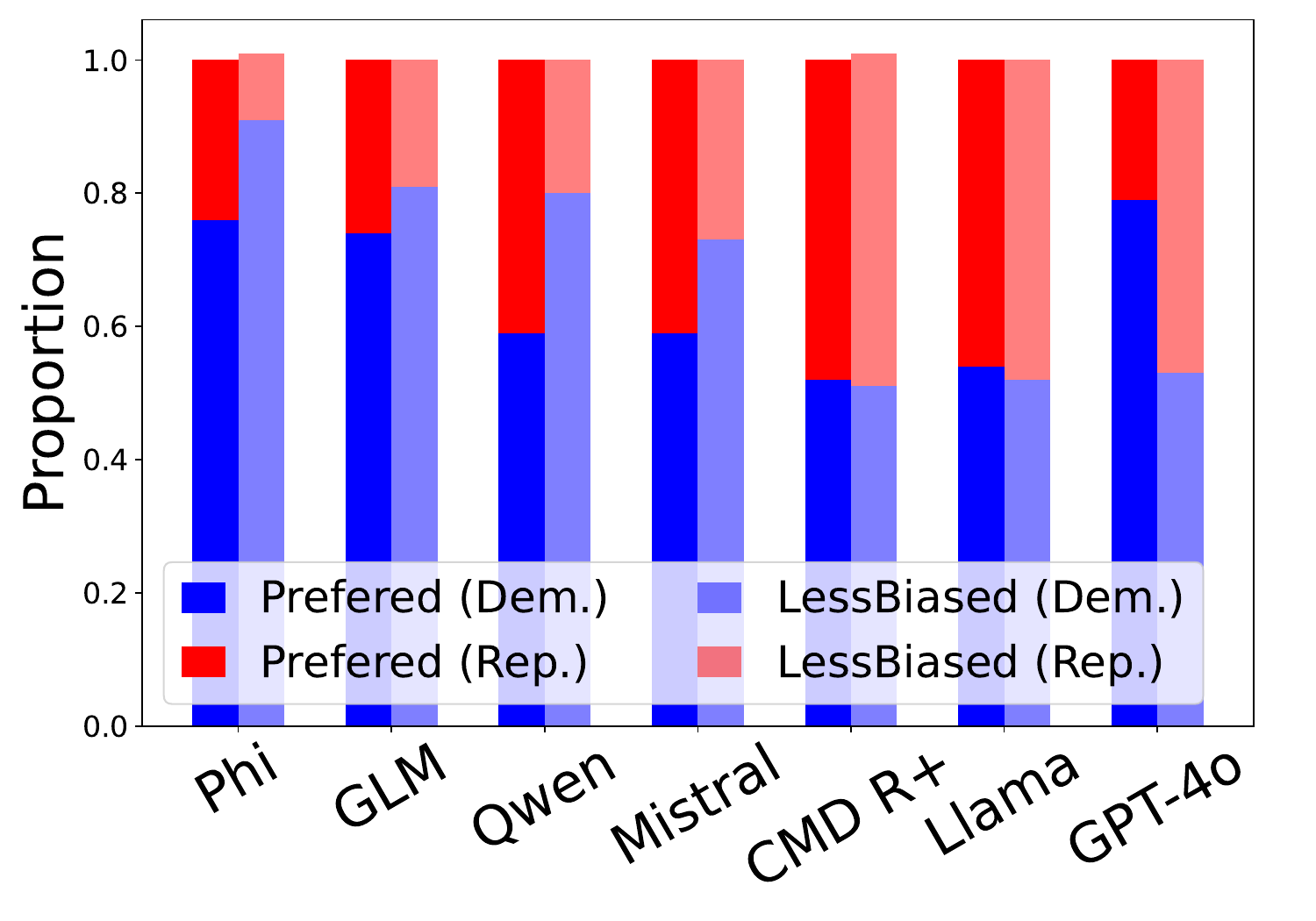}
        \label{fig:5} 
    \end{minipage}
    \hfill
    \begin{minipage}[b]{0.49\linewidth}
        \centering
        \includegraphics[width=\linewidth]{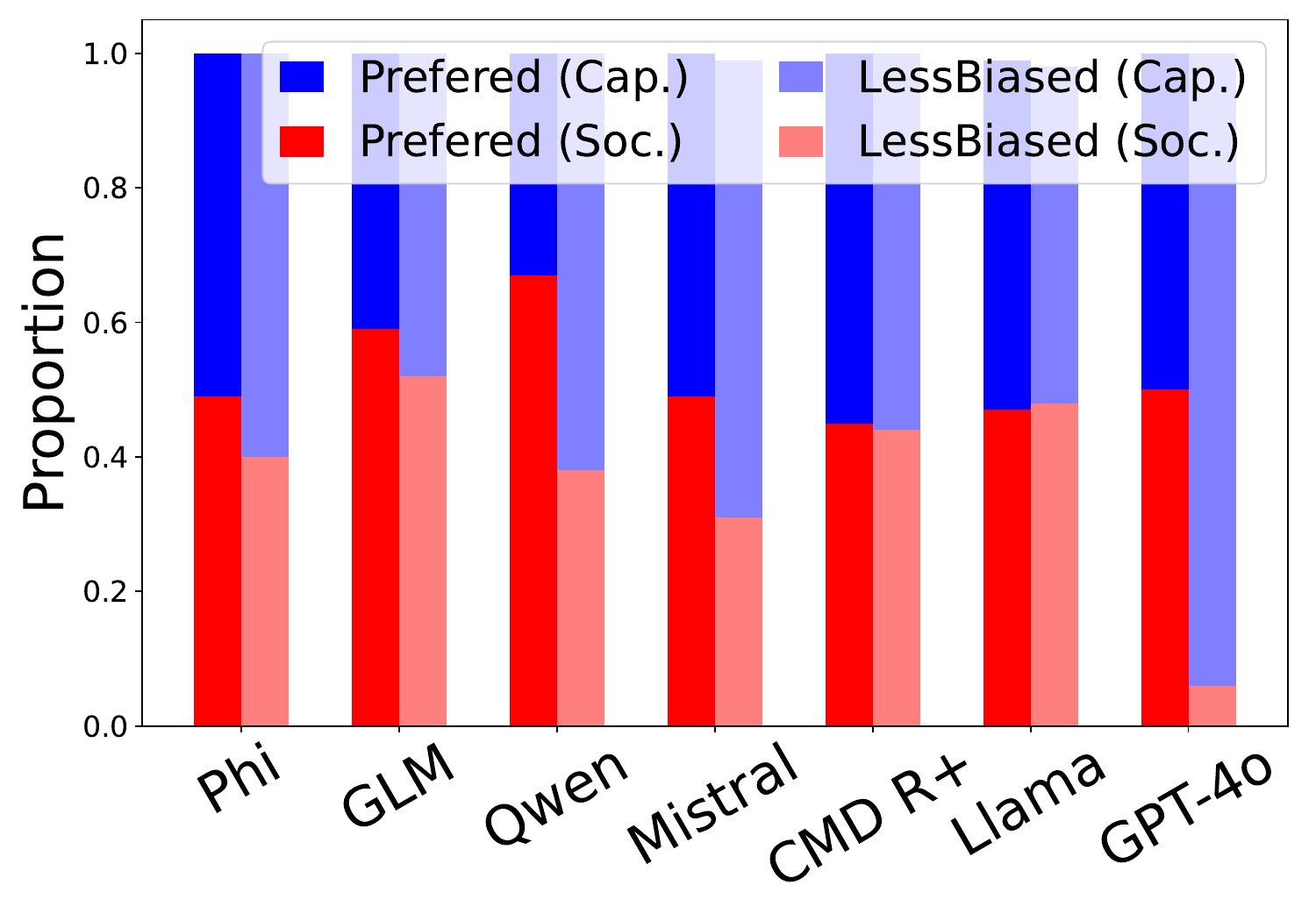}
        \label{fig:6} 
    \end{minipage}

    \caption{%
        Comparison of LLMs’ behavior when asked to choose
        their preferred article
        (prompt 1 in Table 8)
        versus when asked to identify which article is less
        biased
        (prompt 3 in Table 8).
        \textbf{Left:} Political topics. 
        \textbf{Right:}
        Economic topics.}
    \label{fig:prefer_bias}
\end{figure}

We further examine how preference-indication prompts affect model choices.
We compare outcomes under two criteria: \emph{preferred article} and \emph{least-bias} selection (rows~1 and~3 in \Cref{table:pereference_prompts}), with results shown in \Cref{fig:prefer_bias}.
CMD R+ shows the strongest alignment between the two criteria, while GPT-4o diverges notably on \textsc{EconoLex}.
Qwen and Mistral also exhibit substantial gaps.
Models are generally more consistent on political topics than on economic ones, where discrepancies widen.

These results indicate that even top-performing models struggle to generalize across prompt types, highlighting limits to their robustness in evaluative settings.

\subsection{User vs. System Bias}

To analyze media bias in LLMs, we perform an ablation at both the system and user levels.
We define \emph{System Bias} as ideological leaning induced by a biased system prompt (third row in Tables~\ref{table:article_prompt_combos} and~\ref{table:article_prompt_combos_real}), and \emph{User Bias} as bias arising when a neutral model responds to a biased user request (first row).
These settings mirror real-world use cases where news agencies employ either domain-specific or general-purpose LLMs.

\begin{figure}[ht]
    \centering
    \begin{minipage}[b]{0.48\linewidth}
        \centering
        \includegraphics[width=\linewidth]{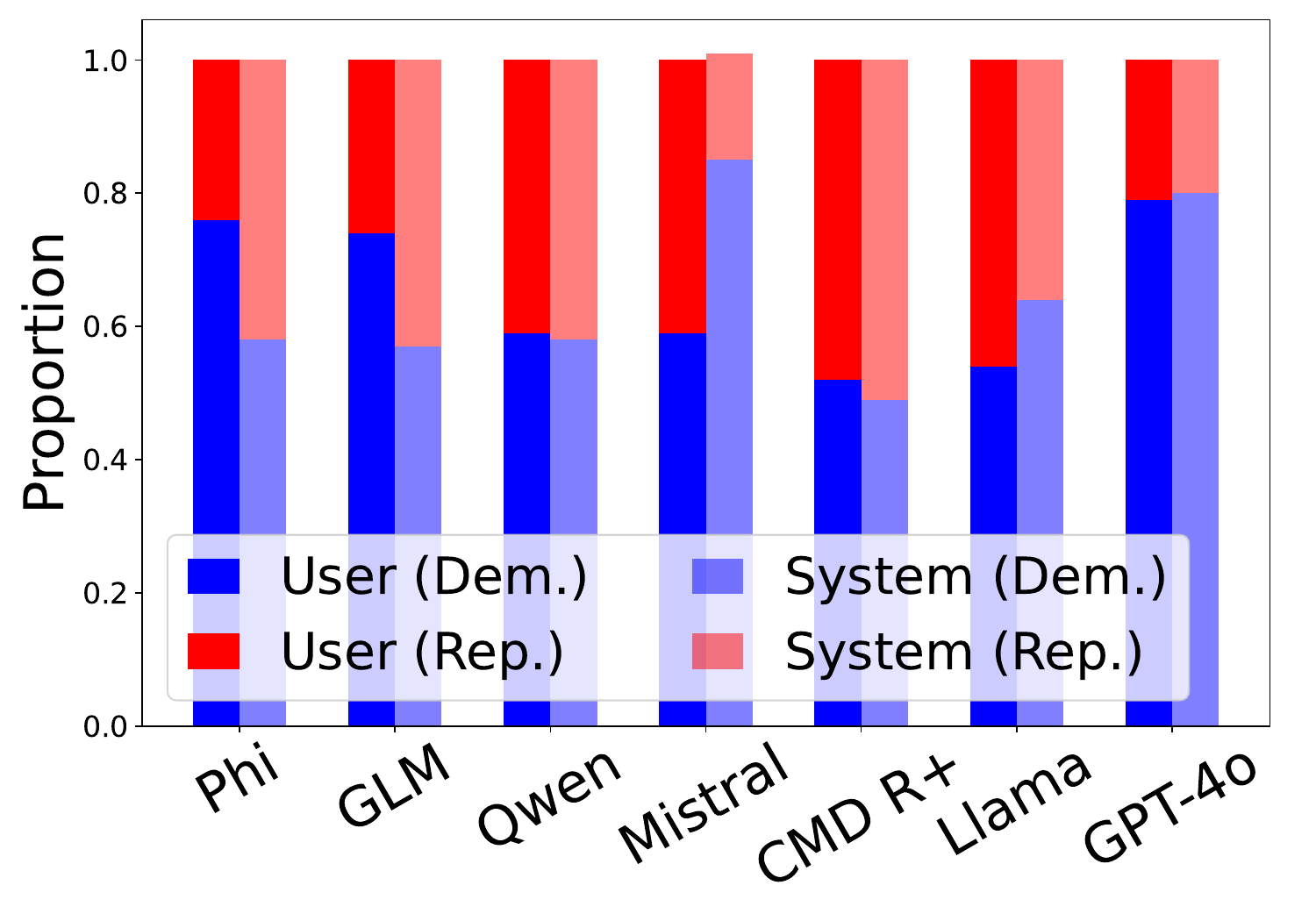}
        \label{fig:3} 
    \end{minipage}
    \hfill
    \begin{minipage}[b]{0.48\linewidth}
        \centering
        \includegraphics[width=\linewidth]{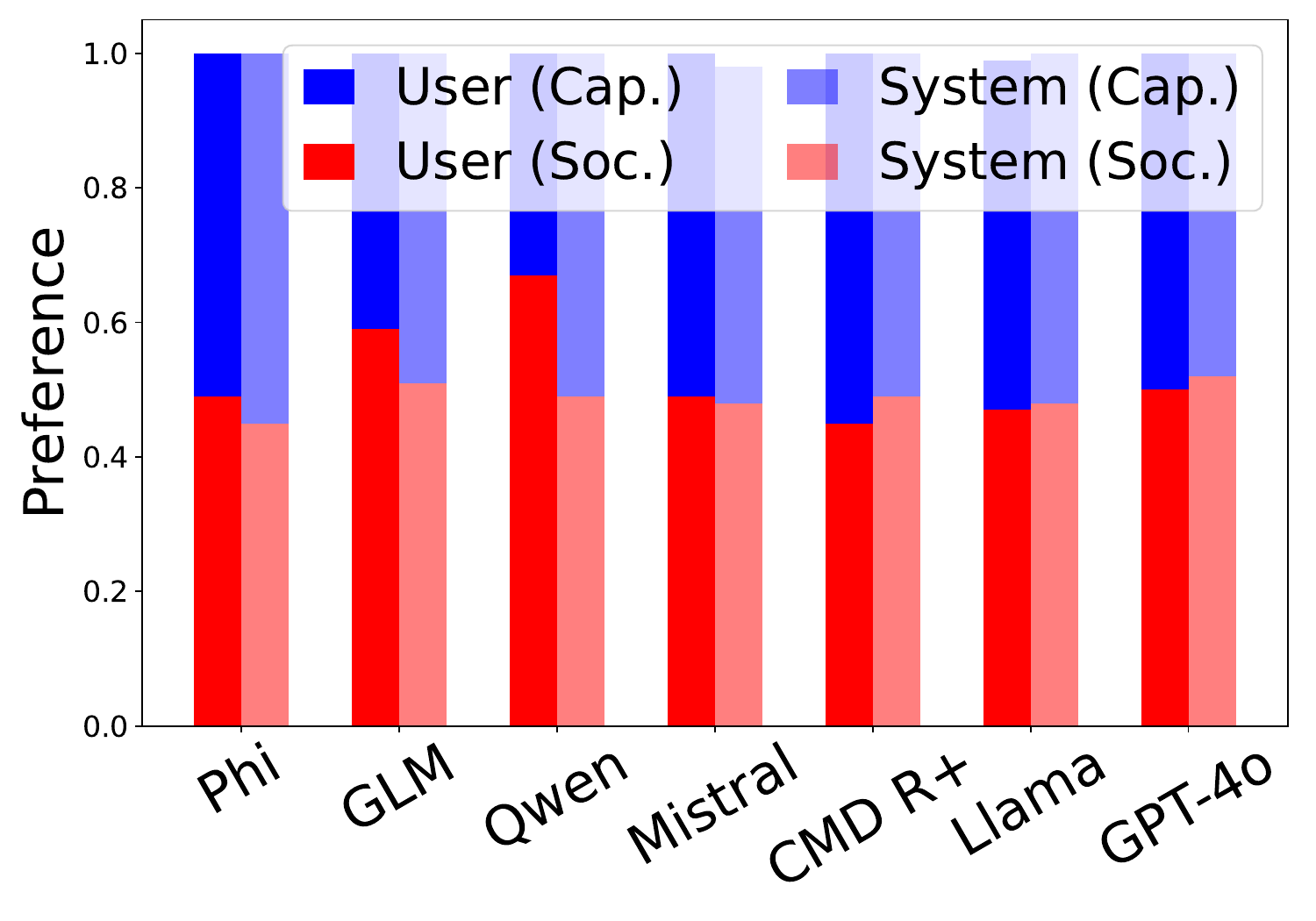}
        \label{fig:4} 
    \end{minipage}

    \caption{Comparison of LLMs’ preferences when the
    article is generated by a biased system
(rows 3 of Tables 2 and 3)
versus a biased user (rows 1 of Tables 2 and 3). 
    \textbf{Left:} Political topics.
    \textbf{Right:} Economic topics. }
    \label{fig:agent_user}
\end{figure}

As shown in \Cref{fig:agent_user}, explicit system-level bias generally yields smaller asymmetries than comparable user-level bias, except for Mistral and Llama on political topics.
GLM and Phi (political) and GLM and Qwen (economic) show reduced asymmetry under biased system prompts.
Given that role distinctions are largely established during post-training (SFT and RLHF), we hypothesize that broader user-role data increases sensitivity to biased user inputs.
CMD R+ exhibits the least role-dependent variation. However, given its poor accuracy and low agreement with human annotations (Table \ref{tab:human-model-results}), it is unclear whether this behavior reflects strong safety tuning or a limited ability to discriminate role differences.
Overall, fluctuations are smaller for economic topics than for political ones.

\subsection{Cross-Model Comparison}

\begin{table}[t]
\centering
\setlength{\tabcolsep}{5pt}\renewcommand{\arraystretch}{1.05}

\resizebox{\linewidth}{!}{%
\begin{tabular}{@{} c l cc cc cc cc @{}}
\toprule
& & \multicolumn{8}{c}{\textbf{Generator}} \\
\cmidrule(lr){3-10}
& & \multicolumn{2}{c}{\textbf{GPT-4o}} & \multicolumn{2}{c}{\textbf{CMD R+}} &
  \multicolumn{2}{c}{\textbf{Qwen}} & \multicolumn{2}{c}{\textbf{CMD R+}} \\
\cmidrule(lr){3-4}\cmidrule(lr){5-6}\cmidrule(lr){7-8}\cmidrule(lr){9-10}
& \textbf{ } & \textbf{DEM} & \textbf{REP} &
  \textbf{DEM} & \textbf{REP} &
  \textbf{SOC} & \textbf{CAP} &
  \textbf{SOC} & \textbf{CAP} \\
\midrule
\multirow{4}{*}{\rotatebox{90}{\textbf{Annotator}}}
& \textbf{GPT-4o} & 0.79 & 0.21 & 0.44 & 0.56 &  &  &  &  \\
& \textbf{CMD R+} & 0.53 & 0.47 & 0.52 & 0.48 &  &  &  &  \\
& \textbf{Qwen}   &      &      &      &      & 0.67 & 0.33 & 0.52 & 0.48 \\
& \textbf{CMD R+} &      &      &      &      & 0.50 & 0.50 & 0.45 & 0.55 \\
\bottomrule
\end{tabular}}
\caption{Cross-model evaluation results on \textsc{PoliGen} (left) and \textsc{EconoLex} (right). The table contrasts each model’s judgments with its own self-annotation results Table~\ref{tab:comparison}, showing that cross-model evaluation reduces the stronger biases observed under self-evaluation.}
\label{tab:crossmodel}
\end{table}

To better isolate ideological asymmetry signals, we conducted cross-model evaluations using the two models that showed the most contrasting preference patterns within each dataset. 
This design tests whether these asymmetries persist when models assess outputs produced by models with opposing preference tendencies (Table~\ref{tab:crossmodel}).

\paragraph{\textsc{PoliGen} Dataset}

In the political domain, GPT-4o showed the strongest Democratic preference under self-evaluation (79\%), while CMD R+ demonstrated the strongest Republican preference (48\%). When these two models evaluated each other’s outputs, their preferences shifted toward balance. GPT-4o favored Republican articles 56\% of the time when judging CMD R+’s texts, whereas CMD R+ expressed a slight Democratic preference (53\%) when evaluating GPT-4o’s outputs (Table~\ref{tab:crossmodel}).

This attenuation of bias is likely due to the interaction between generation and evaluation. Models tend to produce articles that are more convincingly written from the ideological perspective they internally favor, as stylistic and rhetorical choices align with their preferred leaning. However, when those same articles are evaluated by a model with an opposing ideological bias, the persuasive force of the original leaning is reduced, leading to more balanced preference distributions. 

\paragraph{\textsc{EconoLex} Dataset}

In the economic domain, Qwen displayed the strongest socialist preference (67\%), while CMD R+ leaned most toward capitalist articles (55\%). Under cross-model evaluation, both models produced nearly balanced outcomes. Qwen showed a slight preference for capitalist articles (52\%) when judging CMD R+’s texts, while CMD R+’s evaluations of Qwen’s outputs were evenly split between socialist and capitalist perspectives (50\% each) (Table~\ref{tab:crossmodel}).

As in the political domain, this reduction in bias can be explained by the fact that self-aligned outputs are typically optimized toward a model’s favored ideological perspective. When such outputs are instead judged by a model with an opposing leaning, the stylistic and rhetorical advantages of alignment are less persuasive, yielding more balanced outcomes.

\section{Conclusion and Future Work}
We conducted a systematic analysis of ideological framing bias in LLM-generated text across political and economic domains, introducing two new datasets, \textsc{PoliGen} and \textsc{EconoLex}, and evaluating eight prominent models through generation, annotation, and Socratic probing. Our results show that GPT-4o achieves near-human accuracy and agreement in detecting ideological framing, while open-weight models display moderate reliability. Socratic probing reveals that despite high annotation accuracy, LLMs exhibit consistent directional preferences in binary comparisons, GPT-4o favoring capitalist and Democratic framings, and Chinese-developed models favoring socialist perspectives. 
While LLMs identify framing bias effectively, their ideological preferences highlight the need for combined human and model evaluation.

One line for future work is to examine the origins of asymmetries, whether rooted in pre-training data, supervised fine tuning, or alignment strategies like RLHF and RLAIF. Extending the study to diverse domains and multilingual models could clarify cross-cultural variation, as media framing conventions differ markedly across regions \cite{Spinde2023MediaBias}. Developing evaluation protocols that account for these cultural and methodological factors will be essential for creating LLMs that support balanced, interpretable, and globally robust content generation and assessment.

\section{Limitations}

Our study operationalizes model behavior via \emph{preference} and \emph{least-bias} indications, emphasizing outcome patterns rather than the underlying explanations driving model decisions. 
Moreover, experiments were limited to political and economic domains, so results may not generalize to others. 
Finally, while GPT-4o showed high agreement with human annotators, our Socratic probing surfaced systematic irregularities, underscoring the limitations of relying solely on model self-evaluation.

\clearpage

\bibliography{anthology1,custom}

@inproceedings{bender2021stochastic,
  author    = {Emily M. Bender and Timnit Gebru and Angelina McMillan-Major and Shmargaret Shmitchell},
  title     = {On the Dangers of Stochastic Parrots: Can Language Models Be Too Big?},
  booktitle = {Proceedings of the 2021 ACM ConferencThinking more wiselye on Fairness, Accountability, and Transparency (FAccT)},
  pages     = {610--623},
  year      = {2021},
  publisher = {Association for Computing Machinery},
  doi       = {10.1145/3442188.3445922}
}

@article{Rishi2021Foundation,
  author       = {Rishi Bommasani and
                  Drew A. Hudson and
                  Ehsan Adeli and
                  Russ B. Altman and
                  Simran Arora and
                  Sydney von Arx and
                  Michael S. Bernstein and
                  Jeannette Bohg and
                  Antoine Bosselut and
                  Emma Brunskill and
                  Erik Brynjolfsson and
                  Shyamal Buch and
                  Dallas Card and
                  Rodrigo Castellon and
                  Niladri S. Chatterji and
                  Annie S. Chen and
                  Kathleen Creel and
                  Jared Quincy Davis and
                  Dorottya Demszky and
                  Chris Donahue and
                  Moussa Doumbouya and
                  Esin Durmus and
                  Stefano Ermon and
                  John Etchemendy and
                  Kawin Ethayarajh and
                  Li Fei{-}Fei and
                  Chelsea Finn and
                  Trevor Gale and
                  Lauren E. Gillespie and
                  Karan Goel and
                  Noah D. Goodman and
                  Shelby Grossman and
                  Neel Guha and
                  Tatsunori Hashimoto and
                  Peter Henderson and
                  John Hewitt and
                  Daniel E. Ho and
                  Jenny Hong and
                  Kyle Hsu and
                  Jing Huang and
                  Thomas Icard and
                  Saahil Jain and
                  Dan Jurafsky and
                  Pratyusha Kalluri and
                  Siddharth Karamcheti and
                  Geoff Keeling and
                  Fereshte Khani and
                  Omar Khattab and
                  Pang Wei Koh and
                  Mark S. Krass and
                  Ranjay Krishna and
                  Rohith Kuditipudi and
                  et al.},
  title        = {On the Opportunities and Risks of Foundation Models},
  journal      = {CoRR},
  volume       = {abs/2108.07258},
  year         = {2021},
  url          = {https://arxiv.org/abs/2108.07258},
  eprinttype    = {arXiv},
  eprint       = {2108.07258},
  bibsource    = {dblp computer science bibliography, https://dblp.org}
}

@inproceedings{weidinger2022taxonomy,
title = "Taxonomy of risks posed by language models",
author = "Laura Weidinger and Jonathan Uesato and Maribeth Rauh and Conor Griffin and Po-sen Huang and John Mellor and Amelia Glaese and Myra Cheng and Borja Balle and Atoosa Kasirzadeh and Courtney Biles and Sasha Brown and Zac Kenton and Will Hawkins and Tom Stepleton and Abeba Birhane and Hendricks, {Lisa Anne} and Laura Rimell and William Isaac and Julia Haas and Sean Legassick and Geoffrey Irving and Iason Gabriel",
note = "/",
year = "2022",
month = sep,
doi = "10.1145/3531146.3533088",
language = "English",
isbn = "9781450393522",
pages = "214--229",
booktitle = "FAccT '22",
publisher = "ACM",

}

@inproceedings{bang-etal-2024-measuring,
    title = "Measuring Political Bias in Large Language Models: What Is Said and How It Is Said",
    author = "Bang, Yejin  and
      Chen, Delong  and
      Lee, Nayeon  and
      Fung, Pascale",
    editor = "Ku, Lun-Wei  and
      Martins, Andre  and
      Srikumar, Vivek",
    booktitle = "Proceedings of the 62nd Annual Meeting of the Association for Computational Linguistics (Volume 1: Long Papers)",
    month = aug,
    year = "2024",
    address = "Bangkok, Thailand",
    publisher = "Association for Computational Linguistics",
    url = "https://aclanthology.org/2024.acl-long.600/",
    doi = "10.18653/v1/2024.acl-long.600",
    pages = "11142--11159"
}

@inproceedings{lin-etal-2024-indivec,
    title = "{I}ndi{V}ec: An Exploration of Leveraging Large Language Models for Media Bias Detection with Fine-Grained Bias Indicators",
    author = "Lin, Luyang  and
      Wang, Lingzhi  and
      Zhao, Xiaoyan  and
      Li, Jing  and
      Wong, Kam-Fai",
    editor = "Graham, Yvette  and
      Purver, Matthew",
    booktitle = "Findings of the Association for Computational Linguistics: EACL 2024",
    month = mar,
    year = "2024",
    address = "St. Julian{'}s, Malta",
    publisher = "Association for Computational Linguistics",
    url = "https://aclanthology.org/2024.findings-eacl.70/",
    pages = "1038--1050"
}

@inproceedings{trhlik-stenetorp-2024-quantifying,
    title = "Quantifying Generative Media Bias with a Corpus of Real-world and Generated News Articles",
    author = "Trhl{\'i}k, Filip  and
      Stenetorp, Pontus",
    editor = "Al-Onaizan, Yaser  and
      Bansal, Mohit  and
      Chen, Yun-Nung",
    booktitle = "Findings of the Association for Computational Linguistics: EMNLP 2024",
    month = nov,
    year = "2024",
    address = "Miami, Florida, USA",
    publisher = "Association for Computational Linguistics",
    url = "https://aclanthology.org/2024.findings-emnlp.255/",
    doi = "10.18653/v1/2024.findings-emnlp.255",
    pages = "4420--4445"
}

@misc{hernandes2024llmsleftrightcenter,
      title={LLMs left, right, and center: Assessing GPT's capabilities to label political bias from web domains}, 
      author={Raphael Hernandes and Giulio Corsi},
      year={2024},
      eprint={2407.14344},
      archivePrefix={arXiv},
      primaryClass={cs.CL},
      url={https://arxiv.org/abs/2407.14344}, 
}

@inproceedings{Horych2025,
title = {The Promises and Pitfalls of LLM Annotations in Dataset Labeling: a Case Study on Media Bias Detection},
author = {Tomas Horych and Christoph Mandl and Terry Ruas and Andre Greiner-Petter and Bela Gipp and Akiko Aizawa and Timo Spinde},
url = {https://media-bias-research.org/wp-content/uploads/2025/01/Horych2025.pdf},
year = {2025},
date = {2025-01-01},
urldate = {2025-01-01},
booktitle = {Findings of the 2025 Conference of the The Nations of the Americas Chapter of the Association for Computational Linguistics: NAACL 2025},
publisher = {Association for Computational Linguistics},
address = {Albuquerque, USA},
pubstate = {published},
tppubtype = {inproceedings}
}

@article{Spinde2023MediaBias,
title = {The Media Bias Taxonomy: A Systematic Literature Review on the Forms and Automated Detection of Media Bias},
author = {Timo Spinde and Smi Hinterreiter and Fabian Haak and Terry Ruas and Helge Giese and Norman Meuschke and Bela Gipp},
url = {https://media-bias-research.org/wp-content/uploads/2023/12/spinde2023.pdf},
year = {2023},
date = {2023-01-01},
urldate = {2023-01-01},
journal = {arXiv preprint},
pubstate = {published},
tppubtype = {article}
}

@article{motoki2024more,
  title={More human than human: measuring ChatGPT political bias},
  author={Motoki, Fabio and Pinho Neto, Valdemar and Rodrigues, Victor},
  journal={Public Choice},
  volume={198},
  number={1},
  pages={3--23},
  year={2024},
doi= {https://doi.org/10.1007/s11127-023-01097-2},
  publisher={Springer}
}

@misc{instructed2023bias,
  title={Instructed to Bias: Instruction-Tuned Language Models Exhibit Emergent Cognitive Bias}, 
      author={Itay Itzhak and Gabriel Stanovsky and Nir Rosenfeld and Yonatan Belinkov},
      year={2024},
      eprint={2308.00225},
      archivePrefix={arXiv},
      primaryClass={cs.AI},
      url={https://arxiv.org/abs/2308.00225}, 
}

@inproceedings{cho2020investigating,
 author = {Vig, Jesse and Gehrmann, Sebastian and Belinkov, Yonatan and Qian, Sharon and Nevo, Daniel and Singer, Yaron and Shieber, Stuart},
 booktitle = {Advances in Neural Information Processing Systems},
 editor = {H. Larochelle and M. Ranzato and R. Hadsell and M.F. Balcan and H. Lin},
 pages = {12388--12401},
 publisher = {Curran Associates, Inc.},
 title = {Investigating Gender Bias in Language Models Using Causal Mediation Analysis},
 url = {https://proceedings.neurips.cc/paper_files/paper/2020/file/92650b2e92217715fe312e6fa7b90d82-Paper.pdf},
 volume = {33},
 year = {2020}
}

@inproceedings{persistent2021bias,
author = {Abid, Abubakar and Farooqi, Maheen and Zou, James},
title = {Persistent Anti-Muslim Bias in Large Language Models},
year = {2021},
isbn = {9781450384735},
publisher = {Association for Computing Machinery},
address = {New York, NY, USA},
url = {https://doi.org/10.1145/3461702.3462624},
doi = {10.1145/3461702.3462624},
booktitle = {Proceedings of the 2021 AAAI/ACM Conference on AI, Ethics, and Society},
pages = {298–306},
numpages = {9},
location = {Virtual Event, USA},
series = {AIES '21}
}

@misc{beyond2024labelbias,
      title={Beyond Performance: Quantifying and Mitigating Label Bias in LLMs}, 
      author={Yuval Reif and Roy Schwartz},
      year={2024},
      eprint={2405.02743},
      archivePrefix={arXiv},
      primaryClass={cs.CL},
      url={https://arxiv.org/abs/2405.02743}, 
}

@inproceedings{winoqueer2023benchmark,
    title = "{W}ino{Q}ueer: A Community-in-the-Loop Benchmark for Anti-{LGBTQ}+ Bias in Large Language Models",
    author = "Felkner, Virginia  and
      Chang, Ho-Chun Herbert  and
      Jang, Eugene  and
      May, Jonathan",
    editor = "Rogers, Anna  and
      Boyd-Graber, Jordan  and
      Okazaki, Naoaki",
    booktitle = "Proceedings of the 61st Annual Meeting of the Association for Computational Linguistics (Volume 1: Long Papers)",
    month = jul,
    year = "2023",
    address = "Toronto, Canada",
    publisher = "Association for Computational Linguistics",
    url = "https://aclanthology.org/2023.acl-long.507/",
    doi = "10.18653/v1/2023.acl-long.507",
    pages = "9126--9140"
}

@inproceedings{bold2024biasdataset,
  author = {Dhamala, Jwala and Sun, Tony and Kumar, Varun and Krishna, Satyapriya and Pruksachatkun, Yada and Chang, Kai-Wei and Gupta, Rahul},
  title = {BOLD: Dataset and metrics for measuring biases in open-ended language generation},
  booktitle = {FAccT},
  year = {2021}
}

@article{spinde2022neural,
  title={Neural Media Bias Detection Using Distant Supervision With BABE--Bias Annotations By Experts},
  author={Spinde, Timo and Plank, Manuel and Krieger, Jan-David and Ruas, Terry and Gipp, Bela and Aizawa, Akiko},
  journal={arXiv preprint arXiv:2209.14557},
  year={2022}
}

@inproceedings{dong2024fnspid,
  title={Fnspid: A comprehensive financial news dataset in time series},
  author={Dong, Zihan and Fan, Xinyu and Peng, Zhiyuan},
  booktitle={Proceedings of the 30th ACM SIGKDD Conference on Knowledge Discovery and Data Mining},
  pages={4918--4927},
  year={2024}
}

@misc{buyl2025largelanguagemodelsreflect,
      title={Large Language Models Reflect the Ideology of their Creators}, 
      author={Maarten Buyl and Alexander Rogiers and Sander Noels and Guillaume Bied and Iris Dominguez-Catena and Edith Heiter and Iman Johary and Alexandru-Cristian Mara and Raphaël Romero and Jefrey Lijffijt and Tijl De Bie},
      year={2025},
      eprint={2410.18417},
      archivePrefix={arXiv},
      primaryClass={cs.CL},
      url={https://arxiv.org/abs/2410.18417}, 
}

@article{vallejo2023connecting,
  title={Connecting the dots in news analysis: bridging the cross-disciplinary disparities in media bias and framing},
  author={Vallejo, Gisela and Baldwin, Timothy and Frermann, Lea},
  journal={arXiv preprint arXiv:2309.08069},
  year={2023}
}

@article{entman2007framing,
  title={Framing bias: Media in the distribution of power},
  author={Entman, Robert M},
  journal={Journal of communication},
  volume={57},
  number={1},
  pages={163--173},
  year={2007},
  publisher={Oxford University Press}
}

@article{mcquail2020mcquail,
  title={McQuail's media and mass communication theory},
  author={McQuail, Denis and Deuze, Mark},
  year={2020},
  publisher={SAGE Publications Ltd}
}

@INPROCEEDINGS{spinde_thinkIsBiased,
  author={Spinde, Timo and Kreuter, Christina and Gaissmaier, Wolfgang and Hamborg, Felix and Gipp, Bela and Giese, Helge},
  booktitle={2021 ACM/IEEE Joint Conference on Digital Libraries (JCDL)}, 
  title={Do You Think It's Biased? How To Ask For The Perception Of Media Bias}, 
  year={2021},
  volume={},
  number={},
  pages={61-69},
  doi={10.1109/JCDL52503.2021.00018}}

@article{ho2023thinking,
  title={Thinking more wisely: using the Socratic method to develop critical thinking skills amongst healthcare students},
  author={Ho, Yueh-Ren and Chen, Bao-Yu and Li, Chien-Ming},
  journal={BMC medical education},
  volume={23},
  number={1},
  pages={173},
  year={2023},
  publisher={Springer}
}

@article{lin2024investigating,
  title={Investigating bias in llm-based bias detection: Disparities between llms and human perception},
  author={Lin, Luyang and Wang, Lingzhi and Guo, Jinsong and Wong, Kam-Fai},
  journal={arXiv preprint arXiv:2403.14896},
  year={2024}
}

@article{trhlik2024quantifying,
  title={Quantifying generative media bias with a corpus of real-world and generated news articles},
  author={Trhlik, Filip and Stenetorp, Pontus},
  journal={arXiv preprint arXiv:2406.10773},
  year={2024}
}

@article{fang2024bias,
  title={Bias of AI-generated content: an examination of news produced by large language models},
  author={Fang, Xiao and Che, Shangkun and Mao, Minjia and Zhang, Hongzhe and Zhao, Ming and Zhao, Xiaohang},
  journal={Scientific Reports},
  volume={14},
  number={1},
  pages={5224},
  year={2024},
  publisher={Nature Publishing Group UK London}
}

@article{lin2023data,
  title={Data-Augmented and Retrieval-Augmented Context Enrichment in Chinese Media Bias Detection},
  author={Lin, Luyang and Li, Jing and Wong, Kam-Fai},
  journal={arXiv preprint arXiv:2311.01372},
  year={2023}
}

@article{zheng2023judging,
  title={Judging llm-as-a-judge with mt-bench and chatbot arena},
  author={Zheng, Lianmin and Chiang, Wei-Lin and Sheng, Ying and Zhuang, Siyuan and Wu, Zhanghao and Zhuang, Yonghao and Lin, Zi and Li, Zhuohan and Li, Dacheng and Xing, Eric and others},
  journal={Advances in neural information processing systems},
  volume={36},
  pages={46595--46623},
  year={2023}
}

@article{liu2023g,
  title={G-eval: NLG evaluation using gpt-4 with better human alignment},
  author={Liu, Yang and Iter, Dan and Xu, Yichong and Wang, Shuohang and Xu, Ruochen and Zhu, Chenguang},
  journal={arXiv preprint arXiv:2303.16634},
  year={2023}
}

@inproceedings{kim2023prometheus,
  title={Prometheus: Inducing fine-grained evaluation capability in language models},
  author={Kim, Seungone and Shin, Jamin and Cho, Yejin and Jang, Joel and Longpre, Shayne and Lee, Hwaran and Yun, Sangdoo and Shin, Seongjin and Kim, Sungdong and Thorne, James and others},
  booktitle={The Twelfth International Conference on Learning Representations},
  year={2023}
}

@article{kim2024prometheus,
  title={Prometheus 2: An open source language model specialized in evaluating other language models},
  author={Kim, Seungone and Suk, Juyoung and Longpre, Shayne and Lin, Bill Yuchen and Shin, Jamin and Welleck, Sean and Neubig, Graham and Lee, Moontae and Lee, Kyungjae and Seo, Minjoon},
  journal={arXiv preprint arXiv:2405.01535},
  year={2024}
}

@article{gu2024survey,
  title={A survey on llm-as-a-judge},
  author={Gu, Jiawei and Jiang, Xuhui and Shi, Zhichao and Tan, Hexiang and Zhai, Xuehao and Xu, Chengjin and Li, Wei and Shen, Yinghan and Ma, Shengjie and Liu, Honghao and others},
  journal={arXiv preprint arXiv:2411.15594},
  year={2024}
}

@article{potter2024hidden,
  title={Hidden Persuaders: LLMs' Political Leaning and Their Influence on Voters},
  author={Potter, Yujin and Lai, Shiyang and Kim, Junsol and Evans, James and Song, Dawn},
  journal={arXiv preprint arXiv:2410.24190},
  year={2024}
}

@article{theartofsocraticquestioning2023,
  title={The Art of Socratic Questioning: Recursive Thinking with Large Language Models},
  author={Qi, Lei and Xu, Hang and Xu, Yichong and Shen, Zheyan and Liu, Shizhe and Wang, Haibin and Jin, Hanwang},
  journal={arXiv preprint arXiv:2305.14999},
  year={2023}
}

@article{he_selfcorrection2024,
  title={Self-Correction is More than Refinement: A Learning Framework for Visual and Language Reasoning Tasks},
  author={He, Yucheng and Lin, Xiao and Wang, Jiapeng and Fung, Pascale and Ji, Heng},
  journal={arXiv preprint arXiv:2410.04055},
  year={2024}
}

@article{socraticquestioningmultimodal2025,
  title={Socratic Questioning: Learn to Self-guide Multimodal Reasoning in the Wild},
  author={Hu, Tianyu and Liu, Qingyu and Chen, Yijia and Zhou, Yuchen and Zhang, Wenqi and Yu, Zhou},
  journal={arXiv preprint arXiv:2501.02964},
  year={2025}
}

@article{wang2024socreval,
  title={SocREval: Large Language Models with the Socratic Method for Robust Reasoning Evaluation},
  author={Wang, Zonghan and Xiong, Wei and Zhao, Yilun and Wang, Xiaozhi and Diao, Shizhe and Gao, Tianyu},
  journal={arXiv preprint arXiv:2310.00074},
  year={2024}
}

@article{socraticRL2025,
  title={Socratic RL: A Novel Framework for Efficient Knowledge Acquisition through Iterative Reflection and Viewpoint Distillation},
  author={Wu, Yihan},
  journal={arXiv preprint arXiv:2506.13358},
  year={2025}
}

@inproceedings{mokhberian2020moral,
  title={Moral framing and ideological bias of news},
  author={Mokhberian, Negar and Abeliuk, Andr{\'e}s and Cummings, Patrick and Lerman, Kristina},
  booktitle={International conference on social informatics},
  pages={206--219},
  year={2020},
  organization={Springer}
}

@article{pastorino2024decoding,
  title={Decoding news narratives: A critical analysis of large language models in framing detection},
  author={Pastorino, Valeria and Sivakumar, Jasivan A and Moosavi, Nafise Sadat},
  journal={arXiv preprint arXiv:2402.11621},
  year={2024}
}

@inproceedings{wang2025media,
  title={Media Bias Detector: Designing and Implementing a Tool for Real-Time Selection and Framing Bias Analysis in News Coverage},
  author={Wang, Jenny S and Haider, Samar and Tohidi, Amir and Gupta, Anushkaa and Zhang, Yuxuan and Callison-Burch, Chris and Rothschild, David and Watts, Duncan J},
  booktitle={Proceedings of the 2025 CHI Conference on Human Factors in Computing Systems},
  pages={1--27},
  year={2025}
}

@article{li2024llms,
  title={Llms-as-judges: a comprehensive survey on llm-based evaluation methods},
  author={Li, Haitao and Dong, Qian and Chen, Junjie and Su, Huixue and Zhou, Yujia and Ai, Qingyao and Ye, Ziyi and Liu, Yiqun},
  journal={arXiv preprint arXiv:2412.05579},
  year={2024}
}

@inproceedings{rottger2024political,
  title={Political compass or spinning arrow? towards more meaningful evaluations for values and opinions in large language models},
  author={R{\"o}ttger, Paul and Hofmann, Valentin and Pyatkin, Valentina and Hinck, Musashi and Kirk, Hannah and Sch{\"u}tze, Hinrich and Hovy, Dirk},
  booktitle={Proceedings of the 62nd Annual Meeting of the Association for Computational Linguistics (Volume 1: Long Papers)},
  pages={15295--15311},
  year={2024}
}

@inproceedings{haller2025leveraging,
  title={Leveraging In-Context Learning for Political Bias Testing of LLMs},
  author={Haller, Patrick and Vamvas, Jannis and Sennrich, Rico and J{\"a}ger, Lena Ann},
  booktitle={Proceedings of the 63rd Annual Meeting of the Association for Computational Linguistics (Volume 1: Long Papers)},
  pages={24718--24738},
  year={2025}
}

@article{elbouanani2025analyzing,
  title={Analyzing Political Bias in LLMs via Target-Oriented Sentiment Classification},
  author={Elbouanani, Akram and Dufraisse, Evan and Popescu, Adrian},
  journal={arXiv preprint arXiv:2505.19776},
  year={2025}
}

@inproceedings{maab2024media,
  title={Media bias detection across families of language models},
  author={Maab, Iffat and Marrese-Taylor, Edison and Pad{\'o}, Sebastian and Matsuo, Yutaka},
  booktitle={Proceedings of the 2024 Conference of the North American Chapter of the Association for Computational Linguistics: Human Language Technologies (Volume 1: Long Papers)},
  pages={4083--4098},
  year={2024}
}

@article{faulborn2025only,
  title={Only a Little to the Left: A Theory-grounded Measure of Political Bias in Large Language Models},
  author={Faulborn, Mats and Sen, Indira and Pellert, Max and Spitz, Andreas and Garcia, David},
  journal={arXiv preprint arXiv:2503.16148},
  year={2025}
}

@article{gu2025alignment,
  title={Alignment Revisited: Are Large Language Models Consistent in Stated and Revealed Preferences?},
  author={Gu, Zhuojun and Wang, Quan and Han, Shuchu},
  journal={arXiv preprint arXiv:2506.00751},
  year={2025}
}

\appendix

\clearpage

\section{Appendix}
\label{sec:appendix}

\paragraph{Clarification on System Prompt Terminology}
In this paper, 'system prompt' refers to the prompt configurable by the entity hosting the language model, not the internal system prompt set by the model’s developer.

\vspace{1em}

Table \ref{tab:dataset_categories} provides a detailed breakdown of the \textsc{PoliGen} and \textsc{EconoLex} datasets.

\begin{table}[h]
    \centering
    \scriptsize
    \begin{tabular}{llc}
        \toprule
        \textbf{Dataset} & \textbf{Category} & \textbf{Number of Items} \\
        \midrule
        \textbf{PoliGen} & Economic and Financial Issues & 140\\
        & Education and Research & 40\\
        & Environmental and Energy Concerns & 120 \\
        & Government and Legal Systems & 40\\
        & Healthcare and Public Health & 140\\
        & Infrastructure and Development & 120\\
        & National Security and Foreign Relations & 120\\
        & Social Justice and Civil Rights & 40\\
        & Technology and Innovation & 120\\
        & Cultural and Community & 120\\
        \midrule
        \textbf{EconoLex} & Miscellaneous & 563\\
        & Business and Economy & 193\\
        & Energy and Environment & 87\\
        & Health and Medicine & 51\\
        & Politics & 51\\
        & Science \& Technology & 37\\
        & International Affairs & 30\\
        & Social Issues & 15\\
        & Science and Research & 12\\
        & Sports and Entertainment & 7\\
        \bottomrule
    \end{tabular}
    \caption{Categories and Number of Items per category for \textsc{PoliGen} and \textsc{EconoLex} Datasets}
    \label{tab:dataset_categories}
\end{table}

Table \ref{table:pereference_prompts} outlines the structured prompt templates used for eliciting preferences and bias assessments from the language model.

\appendix
\section{Annotator Profile and Guidelines}
\label{appendix:annotators}

\paragraph{Human annotation.}
We collected annotations from 13 annotators. Prior to recruiting crowdworkers, the author team, including media-bias experts, annotated model-specific subsets of EconoLex and reconciled disagreements to establish a high-quality reference standard. We then recruited annotators via Prolific, requiring C2-level English proficiency and administering attention checks. Candidates who failed the attention checks or fell below a predefined agreement threshold against the reference subsets were excluded. We ultimately selected the top three Prolific annotators to label EconoLex subsets across models.

The three selected Prolific annotators were all aged 25--34 and were compensated at \$8/hour: (i) a Nigerian woman based in the UK with a graduate degree in Computer Science who self-identified as Democratic-leaning; (ii) a Russian woman based in the UK, a near-native English speaker, with a graduate degree in the Arts who self-identified as centrist; and (iii) a U.S.\ man based in the U.S., a native English speaker, with a graduate degree in Business/Finance who self-identified as Democratic-leaning.

In addition, 10 university-affiliated annotators (all with C2-level English proficiency and at least an MSc; including PhD students, postdoctoral researchers, and faculty) contributed annotations. Annotators received detailed written guidelines defining ideological framing categories in terms of \textbf{tone, values, prioritized stakeholders, and implied solutions/audience} rather than surface lexical cues. To discourage keyword-based labeling, overt ideological terms were masked and replaced with \texttt{BLANK}. For in-person annotation, we held a brief calibration discussion over example triplets and edge cases prior to labeling. Full annotation guidelines are provided below \ref{app:guidelines}.

\section{Annotation Guidelines}
\label{app:guidelines}

\subsection{Task overview}
Each sample contains \textbf{three anonymized articles} on the same topic. Your task is to assign \textbf{one ideological label to each article}.
\begin{itemize}
    \item \textbf{EconoLex (economic framing):} \textit{Socialist}, \textit{Capitalist}, \textit{Neutral}
\end{itemize}
\noindent\textbf{Important constraint:} You must use \textbf{each label exactly once per set} (no repeats within a set).

Articles have been anonymized, and key ideological terms may be replaced with \textbf{BLANK} to minimize surface-level cues. Focus on \textbf{tone, values, priorities, and implied framing}, not specific word choices.

\subsection{File format and workflow}
Each row contains three articles: \textit{Article~1}, \textit{Article~2}, \textit{Article~3}. Articles appear in the text/metadata columns, and labels are entered in the corresponding label columns.
\begin{itemize}
    \item Per row, assign exactly \textbf{one of each label} (no repeats).
\end{itemize}

\subsection{Ideological categories}

\subsubsection{Economic framing (EconoLex)}
\paragraph{Socialist.}
\textbf{Values:} collective well-being, labor rights, social justice, systemic critique.\\
\textbf{Typical framing:} highlights inequality, critiques corporate power, emphasizes worker agency/collective solutions.\\
\textbf{Tone:} empathetic, reform-oriented, anti-corporate or skeptical of market-first logic.

\paragraph{Capitalist.}
\textbf{Values:} markets, private enterprise, economic growth, investor outcomes.\\
\textbf{Typical framing:} emphasizes market efficiency, business confidence, investment, fiscal/monetary impacts on growth.\\
\textbf{Tone:} market-oriented, pro-growth, investor/firm-centered interpretations.

\paragraph{Neutral.}
\textbf{Values:} balanced reporting, factual tone, minimal evaluative language.\\
\textbf{Typical framing:} presents developments without endorsement; uses institutional sources and avoids interpretation.

\begin{table*}[ht]
  \centering
  \small
  \caption{Prompt Types for Preference Indication. Each prompt includes a \texttt{<User>} role followed by a strict instruction on how to respond. 
  A bit of the assistant's content is added to force the model to respond in the expected format using the \texttt{<Assistant>} role tag.}
  \label{preference_prompts}
  \begin{tabular}{c|p{12.5cm}}
    \toprule
    \textbf{nº} & \textbf{Prompt Template} \\
    \midrule

    1 & \textbf{Two-Way Preference Indication}\newline
        \texttt{<User> You will read two articles. Which article do you prefer? Respond strictly with '1' or '2'. <end>}\newline
        \texttt{<Assistant> My preferred article is:}\\[6pt]
    \midrule

    2 & \textbf{Three-Way Preference Indication}\newline
        \texttt{<User> You will read three articles. Which article do you prefer? Respond strictly with '1', '2', or '3'. <end>}\newline
        \texttt{<Assistant> My preferred article is:}\\[6pt]
    \midrule

    3 & \textbf{Two-Way Least-Biased Indication}\newline
        \texttt{<User> Which of these two articles is less biased? Respond strictly with '1' or '2'. <end>}\newline
        \texttt{<Assistant> The least biased article is:}\\[6pt]
    \midrule

    4 & \textbf{Three-Way Least-Biased Indication}\newline
        \texttt{<User> Which of these three articles is less biased? Respond strictly with '1', '2', or '3'. <end>}\newline
        \texttt{<Assistant> The least biased article is:}\\[6pt]




    \bottomrule
  \end{tabular}
  \label{table:pereference_prompts}
\end{table*}

\vspace{1em}
\begin{table}[h]
\centering
\caption{Number of Human-Annotated Samples per Model}
\label{tab:annotation-counts}
\small
\begin{tabular}{@{}lc@{}}
\toprule
\textbf{Model} & \textbf{Annotated Samples} \\
\midrule
C4ai     & 42 \\
Qwen     & 42 \\
Mistral  & 42 \\
GLM      & 42 \\
Gemma    & 42 \\
Phi      & 42 \\
Llama    & 42 \\
\bottomrule
\end{tabular}
\end{table}


\begin{figure}[t]
  \centering
  \fbox{%
    \begin{minipage}{0.95\linewidth}
    \small\ttfamily
You will be shown three articles. Classify each article as: \\
1 = Socialist \\
2 = Capitalist \\
3 = Neutral \\
\\
Article 1: \\
\textit{[ARTICLE\_1\_TEXT]} \\
\\
Article 2: \\
\textit{[ARTICLE\_2\_TEXT]} \\
\\
Article 3: \\
\textit{[ARTICLE\_3\_TEXT]} \\
\\
Respond strictly with three numbers (1, 2, or 3), separated by commas, corresponding to your classifications for Article 1, Article 2, and Article 3. Example: `1, 2, 3'. \\
Your classifications:
    \end{minipage}%
  }
  \caption{Model prompt used to classify each of three articles into Socialist, Capitalist, or Neutral framing.}
  \label{fig:model-prompt}
\end{figure}

\end{document}